\documentclass[letterpaper, 10pt, conference]{ieeeconf}      

\IEEEoverridecommandlockouts                              

\usepackage[left=54pt,right=54pt,top=54pt,bottom=72pt]{geometry}

\usepackage{multirow}
\usepackage{hyperref} 
\usepackage{booktabs}
\usepackage{subcaption}
\usepackage{graphics} 
\usepackage{epsfig} 
\usepackage{times} 
\usepackage{amsmath} 
\usepackage{amssymb}
\usepackage{url}
\let\labelindent\relax
\usepackage{enumitem}

\title{\LARGE \bf
Sliced Wasserstein Discrepancy in Disentangling Representation and Adaptation Networks for Unsupervised Domain Adaptation}

\author{Joel Sol$^{1}$,  Shadi Alijani$^{1}$ and Homayoun Najjaran$^{1}$ 
\thanks{$^{1}$ Faculty of Engineering and Computer Science, University of Victoria, Victoria, BC V8P 5C2, Canada
        {\tt\small \{joelsol, shadialijani, najjaran\}@uvic.ca}}%
\thanks{The authors have provided supplementary material (code) available at \protect\url{https://github.com/JoelESol/DRANet-SWD}}
}

\begin{document}

\maketitle
\thispagestyle{empty}
\pagestyle{empty}

\begin{abstract}
This paper introduces DRANet-SWD as a novel complete pipeline for disentangling content and style representations of images for unsupervised domain adaptation (UDA). The approach builds upon DRANet by incorporating the sliced Wasserstein discrepancy (SWD) as a style loss instead of the traditional Gram matrix loss. The potential advantages of SWD over the Gram matrix loss for capturing style variations in domain adaptation are investigated. Experiments using digit classification datasets and driving scenario segmentation validate the method, demonstrating that DRANet-SWD enhances performance. Results indicate that SWD provides a more robust statistical comparison of feature distributions, leading to better style adaptation. These findings highlight the effectiveness of SWD in refining feature alignment and improving domain adaptation tasks across these benchmarks. Our code can be found 
\href{https://github.com/JoelESol/DRANet-SWD}{here}.
\end{abstract}

\section{Introduction}

Unsupervised domain adaptation (UDA) is a critical challenge in machine learning, where the objective is to bridge the distributional gap between labeled source data and unlabeled target data. Traditional UDA methods primarily rely on domain alignment techniques such as adversarial training, discrepancy minimization, and reconstruction-based strategies. These approaches primarily align distributions without explicitly modeling content and style separately, which can sometimes lead to less effective adaptation.

Recently, UDA research has shifted more-so towards learning feature representations of the dataset distributions. Some existing methods decompose features into shared and domain-exclusive components, often requiring multiple encoders and generators specialized for each domain \cite{liu2018detachadaptlearningcrossdomain} \cite{gonzalezgarcia2018imagetoimagetranslationcrossdomaindisentanglement} \cite{NIPS2016_45fbc6d3}. In contrast, Disentangling Representation and Adaptation Networks (DRANet) \cite{DRANet_CVPR2021} introduced a single encoder-generator architecture that effectively separates content and style latent spaces. DRANet has previously demonstrated state of the art performance across various UDA tasks, including classification and segmentation.

While existing style transfer methods, such as those utilizing Gram matrix-based losses \cite{gatys2016image}, have shown promise in aligning stylistic features, they are inherently limited to capturing second-order statistical correlations between feature activations. This restriction can lead to incomplete style representation, particularly when adapting between domains with complex distributional shifts, synthetic-to-real scenarios. The sliced Wasserstein discrepancy (SWD) \cite{Apple_SWD_CVPR} offers a more principled approach to style alignment by leveraging optimal transport theory to measure and minimize the divergence between entire feature distributions. Building on this foundation, we propose DRANet-SWD, an extension that incorporates SWD for style representation. By leveraging SWD as a style loss, our approach enhances stylistic feature alignment between domains. Unlike the previously used Gram matrix loss, which captures feature correlations, SWD provides a more robust statistical comparison of feature distributions, enabling more precise style adaptation.

The key contributions of this work are as follows:
\begin{itemize}
    \item We introduce DRANet-SWD and demonstrate how SWD improves the disentanglement of style and content representations in UDA.
    \item We conduct a comprehensive evaluation of DRANet-SWD on benchmark datasets, including digit classification tasks with MNIST \cite{MNIST}, MNIST-M \cite{MNISTM}, USPS \cite{USPS}, and SVHN \cite{SVHN} as well as complex segmentation datasets such as GTA5 \cite{GTA5} and CityScapes \cite{Cityscapes}.
\end{itemize}

\section{Background}
Unsupervised Domain Adaptation (UDA) bridges the domain gap between labeled source data and unlabeled target data by aligning their feature distributions \cite{alijani2024vision}. Theoretical foundations of \cite{ben2010theory} established generalization bounds for domain adaptation. Modern UDA methods fall into three categories: \textbf{Discrepancy-based approaches}, which minimize domain divergence via metrics like Maximum Mean Discrepancy (MMD) \cite{long2015learning} or Correlation Alignment (CORAL) \cite{sun2016deep}. \textbf{Adversarial methods}, which employ domain discriminators to align features \cite{ganin2015unsupervised,tzeng2017adversarial}; and \textbf{Reconstruction-based techniques}, which enforce domain invariance through cycle-consistency or feature reconstruction \cite{hoffman2018cycada,ghifary2016deep}. For example, Cycle-Consistent Adversarial Domain Adaptation (CyCADA) \cite{hoffman2018cycada} aligns domains by reconstructing translated images, ensuring semantic consistency across domains.
Recent adversarial UDA innovations include gradient-aligned networks \cite{ran2024gradient} for improved feature alignment and multi-level joint distribution alignment \cite{liu2024multi} to address both marginal and conditional shifts. Symmetric consistency frameworks with cross-domain mixup \cite{cai2024symmetric} further enhance bidirectional adaptation.
\subsection{Style Transfer and Domain Adaptation}
Style transfer, initially designed for artistic image synthesis \cite{gatys2016image}, has become instrumental in UDA for mitigating domain shifts caused by stylistic variations (e.g., lighting, texture). The Gram matrix, which captures feature correlations to represent style, underpins many methods. \cite{huang2017arbitrary} advanced this with Adaptive Instance Normalization (AdaIN) to align feature statistics across domains. In the context of UDA, \cite{dundar2020domain} pioneered style-aware adaptation by using adversarial training and instance normalization for semantic segmentation, demonstrating robustness to domain shifts. Similarly, \cite{hoffman2018cycada} integrated style transfer with cycle-consistent adaptation to preserve semantic content while aligning domains. 

\subsection{Style Losses in UDA}
Derived from neural style transfer, Gram losses align stylistic attributes by matching feature correlations between domains. Gram loss has been used ubiquitously for style and textural loss due to its computational simplicity and its ability to capture textures. Gram loss works by taking feature layers and computing the Gram matrices and computing the mean squared error. While sliced Wasserstein discrepancy requires additional computation but can be more effective as a style loss. Sliced Wasserstein discrepancy works by projecting n-dimensional features onto random direction vectors, sorting the 1D projections and computing the L2 difference between the sorted list. SWD is able to capture complete stationary statistics of deep features unlike Gram loss \cite{heitz2021slicedwassersteinlossneural}. The driving motivation to change the style loss to SWD is that by better capturing the statistics of the features in the perceptual network, the resulting images generated will be better.

\section{Methodology}

\subsection{Disentangling Representation and Adaptation Networks}

\begin{figure*}
    \centering
    \includegraphics[width=0.90\textwidth]{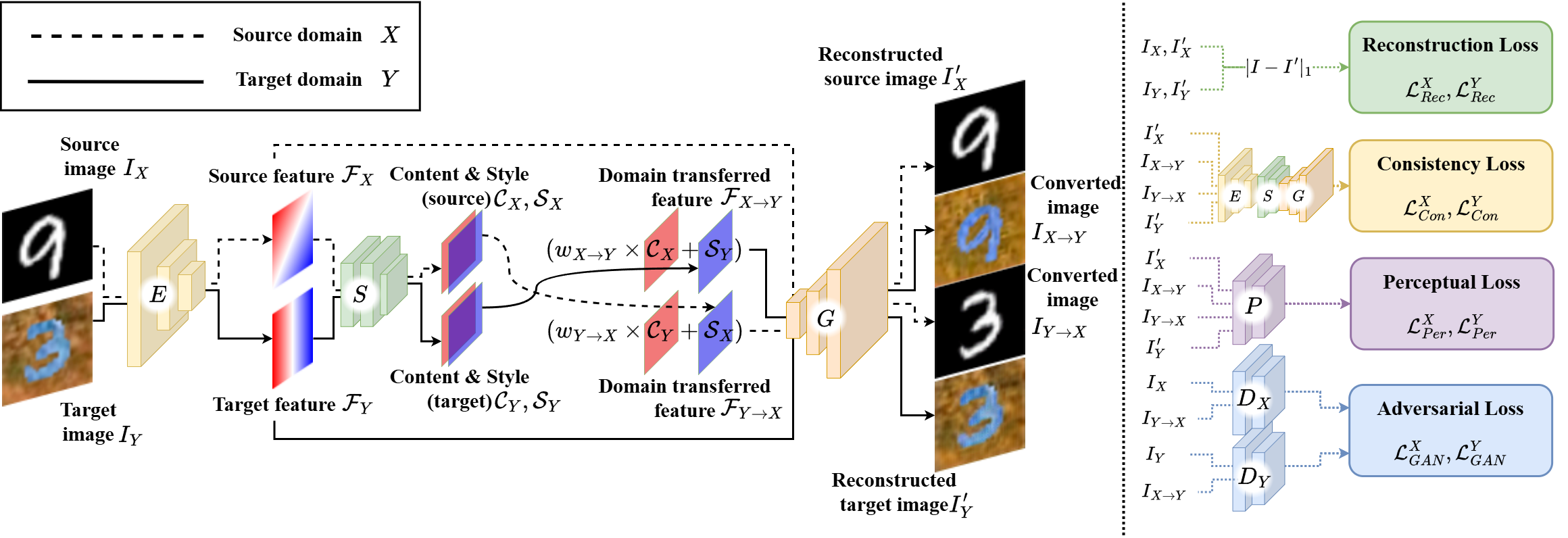}
    \caption{DRANet/DRANet-SWD Architecture}
    \label{CASNet_Architecture}
\end{figure*}

The DRANet-SWD architecture is built off of the previous work from DRANet\cite{DRANet_CVPR2021}. DRANet consists of multiple networks which work to take two or more image datasets and gain an understanding of the image's content and style. DRANet uses these content and style representations in order to adapt the content of one image into the domain of another using the other domain's style representation. The architecture of DRANet is shown in Fig \ref{CASNet_Architecture}. The networks that make up DRANet are an encoder $E$, separator $S$, generator $G$, two discriminators $D_X$, $D_Y$, and a perceptual network $P$.

DRANet uses an image from both domains $I_X$, $I_Y$ where feature spaces $\mathcal{F}_X$, $\mathcal{F}_Y$ are created by the encoder $E$. The feature spaces are used by separator $S$ to be split in contents $\mathcal{C}_X$, $\mathcal{C}_Y$ and styles $\mathcal{S}_X$, $\mathcal{S}_Y$. The styles are then swapped to each domain where $\mathcal{C}_X$ is with $\mathcal{S}_Y$ and $\mathcal{C}_Y$ is with $\mathcal{S}_X$. These content and style representations are then sent to generator $G$. The generator outputs converted images $I_{X\rightarrow Y}$ and $I_{Y\rightarrow X}$. Additionally the generator is used to create reconstructed images $I'_X$, $I'_Y$ by not swapping the style representations and therefore not adapting the images. There is a discriminator for each domain $D_X$, $D_Y$ which analyzes the original and converted images for each domain. For example discriminator $D_X$ analyzes images $I_X$ and $I_{Y\rightarrow X}$. The perceptual network $P$ is used to train the separator $S$ to differentiate content and style.

The perceptual network is used to train the separator network to be able to determine the content and style from the features from the encoder. To do this, the reconstructed $I'_X$, $I'_Y$ and converted images $I_{X\rightarrow Y}$, $I_{Y\rightarrow X}$ were fed into the pre-trained VGG-19 \cite{VGG-19} perceptual network. The outputs from each of the perceptual network layers were used to create the content and style losses. Its important to note that content loss uses features from the last layer of the perceptual network only. As noted by Johnson et al. \cite{DBLP:journals/corr/JohnsonAL16}, using earlier layers in a perceptual network for content loss tend to cause converted images $I_{X\rightarrow Y}$ to be visually indistinguishable from the original image $I_X$. The style losses use the features from all the layers of the perceptual network except for the final layer in order to capture different levels of abstraction and more effectively capture the style in the images. The style loss used in the perceptual loss makes a difference in the resulting generated image quality. The style losses being compared are the Gram matrix style loss, which was used originally in DRANet, and SWD style loss.

\subsection{Tasks and Datasets}
To evaluate the effectiveness of SWD, we consider both classification and segmentation tasks under UDA settings.

For classification, we perform UDA across multiple digit recognition datasets, where the model adapts and classifies between different domains without access to labeled target data. The chosen datasets include MNIST \cite{MNIST}, MNIST-M \cite{MNISTM}, USPS \cite{USPS}, and Street View House Numbers (SVHN) \cite{SVHN}. We evaluate adaptation bi-directionally between MNIST and MNIST-M, MNIST and USPS, as well as MNIST and SVHN.

For segmentation, we assess UDA in a driving scene scenario, where the model adapts from synthetic to real-world images. Specifically, we use the GTA5 dataset \cite{GTA5}, which consists of high-resolution synthetic urban driving scenes, as the source domain, and the Cityscapes dataset \cite{Cityscapes}, which contains real-world street scenes, as the target domain.

\section{Training}

The encoder $E$, separator $S$ and generator $G$ are trained to minimize the overall loss function $\mathcal{L}^d$ while the discriminators $D_d$ try to maximize the loss function. The domain $d$ correspond to the domains $X$ and $Y$.
\begin{equation}
    \label{eq:GAN loss}
    \underset{E,S,G}{\text{min}}\left(\sum_{d\in\{X,Y\}}\underset{Dd}{\text{max}}\mathcal{L}^d\right)
\end{equation}

$\mathcal{L}^d$ consists of multiple components consisting of reconstruction $\mathcal{L}^d_{Rec}$, consistency $\mathcal{L}^d_{Con}$, adversarial $\mathcal{L}^d_{GAN}$ and perceptual $\mathcal{L}^d_{Per}$ losses with each term weighted by a corresponding $\alpha$ term.
\begin{figure*}[htbp]
    \centering
    \begin{subfigure}[b]{0.32\textwidth}
        \centering
        \includegraphics[width=0.75\textwidth]{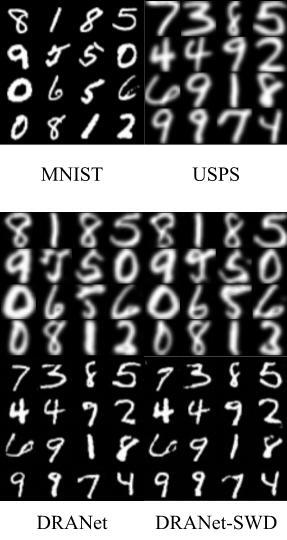}
        \caption{MNIST $\longleftrightarrow$ USPS domain adaptation\\
         \hspace{10mm}}
         \label{fig:digit_a}
    \end{subfigure}
    \begin{subfigure}[b]{0.32\textwidth}
        \centering
        \includegraphics[width=0.75\textwidth]{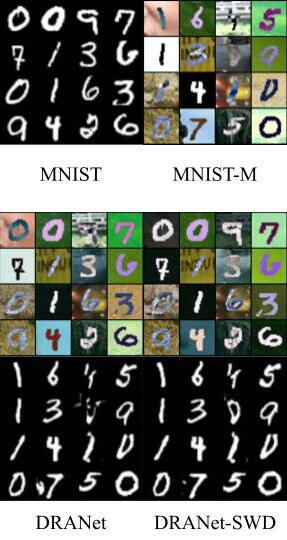}
        \caption{MNIST $\longleftrightarrow$ MNIST-M domain adaptation}
        \label{fig:digit_b}
    \end{subfigure}
    \begin{subfigure}[b]{0.32\textwidth}
        \centering
        \includegraphics[width=0.75\textwidth]{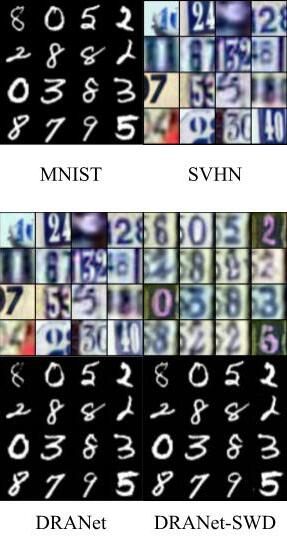}
        \caption{MNIST $\longleftrightarrow$ SVHN domain adaptation\\
         \hspace{10mm}}
         \label{fig:digit_c}
    \end{subfigure}
    \caption{Domain adaptation results for DRANet and DRANet-SWD. The networks use the top row as source/target images. The middle and bottom rows are adaptation results with DRANet left column and DRANet-SWD right column.}
    \label{fig:Digit classification}
\end{figure*}
\begin{equation}
        \label{eq:Domain loss}
        \mathcal{L}^d = \alpha_1\mathcal{L}^d_{Rec} + \alpha_2\mathcal{L}^d_{Con} + \alpha_3\mathcal{L}^d_{GAN}+ \alpha_4\mathcal{L}^d_{Per}
\end{equation}

The losses are discussed in more detail in what follows.

\begin{itemize}[leftmargin=0pt]
    \item[] \textbf{Reconstruction loss}: An L1 loss is used to minimize the difference between input images $I_d$ and reconstructed image $I'_d$ after passing through $E$ and $G$.
    \begin{equation}
        \label{eq:Rec loss}
        \mathcal{L}^d_{Rec} = \mathcal{L}_1(I_d, I'_d), \; \text{where} \; I'_d=G(E(I_d))
    \end{equation}

    \item[] \textbf{Consistency loss}: The consistency loss is used to retain content and style components be comparing feature spaces of the converted and non-converted images from the separator network.
    \begin{equation}
        \label{eq:Con loss}
        \begin{gathered}
            \mathcal{L}^X_{Con} = \mathcal{L}_1(\mathcal{C}_X, \mathcal{C}_{X\rightarrow Y}) + \mathcal{L}_1(\mathcal{S}_X, \mathcal{S}_{Y\rightarrow X}),\\
            \mathcal{L}^Y_{Con}=\mathcal{L}_1(\mathcal{C}_Y, \mathcal{C}_{Y\rightarrow X}) + \mathcal{L}_1(\mathcal{S}_Y, \mathcal{S}_{X\rightarrow Y})
        \end{gathered}
    \end{equation}
    The content $\mathcal{C}_{X\rightarrow Y}$, $\mathcal{C}_{Y\rightarrow X}$ and style $\mathcal{S}_{X\rightarrow Y}$, $\mathcal{S}_{Y\rightarrow X}$ feature spaces are obtained by passing converted images $I_{X\rightarrow Y}$, $I_{Y\rightarrow X}$ through the same encoder $E$ and separator $S$.

    \item[] \textbf{Adversarial loss}: Two discriminators $D_{d\in\{X, Y\}}$ were used for evaluating the adversarial loss for the source and target domains.For the transfer of both domains, the hinge version of adversarial loss is applied and the domain adaptation of $X \rightarrow Y$ is shown below.
    \begin{equation}
        \label{eq:Adversarial loss}
        \begin{gathered}
            \mathcal{L}^d_{GAN}=\mathbb{E}_{y\sim p_{data(Y)}} [\log D_Y(y)] \\
            + \mathbb{E}_{(x,y)\sim p_{data(X,Y)}}[\log (1-D_Y(I_{X\rightarrow Y}(x,y)]
        \end{gathered}
    \end{equation}
    The hinge variation of adversarial loss is used for segmentation tasks.

    \item[] \textbf{Perceptual loss}: The perceptual loss\cite{DBLP:journals/corr/JohnsonAL16} consist of content and style losses and is described in the following equation. A weighting term $\lambda$ is used to balance the content and style.
    \begin{equation}
        \mathcal{L}^d_{Per} = \mathcal{L}^d_{Content} + \lambda\mathcal{L}^d_{Style}
    \end{equation}
    These losses are used to guide the separator network on disentangling the content and style from the features from the encoder. The content loss consists of an MSE loss between feature layers from the perceptual network $P$ from the original and converted images. A specific subset of all the layers $l\in L_C$ of the perceptual network $P$ are used for the content loss. The layers in the perceptual used for content loss $\mathcal{L}^Y_{Content}$ consist of only the final layer of the perceptual network.
    \begin{equation}
        \begin{gathered}
            \mathcal{L}^Y_{Content} = \sum_{l\in L_C}||P_l(I_X) - P_l(I_{X\rightarrow Y})||^2_2\\
            \mathcal{L}^X_{Content} = \sum_{l\in L_C}||P_l(I_Y) - P_l(I_{Y\rightarrow X})||^2_2
        \end{gathered}
    \end{equation}

    The previous style loss used was the Gram matrix style loss described in the following equation.
    \begin{equation}
        \begin{gathered}
        \mathcal{L}^Y_{Style} = \sum_{l\in L_S}||\mathcal{G}(P_l(I_Y)) - \mathcal{G}(P_l(I_{X\rightarrow Y}))||^2_F\\
        \mathcal{L}^X_{Style} = \sum_{l\in L_S}||\mathcal{G}(P_l(I_X)) - \mathcal{G}(P_l(I_{Y\rightarrow X}))||^2_F
        \end{gathered}
    \end{equation}
    A function takes the Gram matrix $\mathcal{G}$ for each layer $l$ form a specific subset of the layers $l\in L_S$ from the perceptual network $P$. The MSE loss is then calculated between the differences in Gram matrices from the original images and the converted images. This was the loss used in the original design of DRANet \cite{DRANet_CVPR2021}. This loss has been replaced with sliced Wasserstein discrepancy, the equation for with is described below.

    \begin{equation}
        \begin{split}
            \mathcal{L}^Y_{Style} = \sum_{l\in L_S} \mathcal{L}_{SW}(P_l(I_Y), P_l(I_{X\rightarrow Y}))\\
            \mathcal{L}^X_{Style} = \sum_{l\in L_S} \mathcal{L}_{SW}(P_l(I_X), P_l(I_{Y\rightarrow X}))
        \end{split}
    \end{equation}
    The Sliced Wasserstein loss $\mathcal{L}_{SW}$ is summed over a subset of layers $l\in L_S$ from the perceptual network $P$ viewing an original image $I_X$, $I_Y$ and converted image $I_{Y\rightarrow X}$, $I_{X\rightarrow Y}$ respectively. The layers $L_S$ consist of all layers of the perceptual network except the final layer. For simplicity $F$ and $F'$ represent a single feature layer from the perceptual network interpreting the original image $I_X$ and converted image $I_{Y\rightarrow X}$. 

    $\mathcal{L}_{SW}(F,F')$ is the Sliced Wasserstein distance between the layer features $F$, $F'$. The features are projected on an n-dimensional unit hypersphere using random directional vectors $V \in \mathcal{S}^{N_l}$. $F_V$, $F'_V$ consist of an unordered set of scalar dot products between the feature vectors and direction $V$.

    \begin{equation}
        \mathcal{L}_{SW}(F, F') = \mathbb{E}_V[\mathcal{L}_{SW1D}(F_V, F'_V)]
    \end{equation}

    $\mathcal{L}_{SW1D}$ is the 1D optimal transport loss between the unordered set of scalars from the directional slices and is defined as the element-wise L2 distance over the sorted lists.
    \begin{equation}
        \mathcal{L}_{SW1D}(S, S')= \dfrac{1}{|S|}||\text{sort}(S)-\text{sort}(S')||^2
    \end{equation}
    
\end{itemize}

For digit classification tasks a batch size of 32 was used and was trained for 40000 iterations using an NVIDIA 4070Ti-Super.
For the GTA5 to Cityscapes segmentation DRANet and DRANet-SWD were trained for 40000 iterations. Additionally, another trial run was performed with additional pretraining of DRN26 \cite{DRN26} for 1700 iterations on just GTA5 using a batch size of 64. DRANet and DRANet-SWD using the pretrained version of DRN26 were trained for 35000 iterations. All segmentation training took place using an NVIDIA A6000 and a batch size of 2.

\section{Results and Discussion}
\subsection{Digit Classification}
\begin{table*}[htpb]
    \centering
    \caption{Result comparison of DRANet and DRANet-SWD on domain adaptation for semantic segmentation}
    \footnotesize
    \setlength{\tabcolsep}{3pt}
    \begin{tabular}{l|c|ccccccccccccccccccc}
        \toprule
        \textbf{Network} & \textbf{mIoU} & \rotatebox{90}{road} & \rotatebox{90}{sidewalk} & \rotatebox{90}{building} & \rotatebox{90}{wall} & \rotatebox{90}{fence} & \rotatebox{90}{pole} & \rotatebox{90}{traffic light} & \rotatebox{90}{traffic sign} & \rotatebox{90}{vegetation} & \rotatebox{90}{terrain} & \rotatebox{90}{sky} & \rotatebox{90}{person} & \rotatebox{90}{rider} & \rotatebox{90}{car} & \rotatebox{90}{truck} & \rotatebox{90}{bus} & \rotatebox{90}{train} & \rotatebox{90}{motorcycle} & \rotatebox{90}{bicycle} \\
        \midrule
        DRANet & 33.63 & 76.8 & 27.1 & 75.9 & \textbf{20.0} & 13.2 & \textbf{22.2} & 23.2 & 11.5 & 80.8 & 33.2 & 76.9 & 45.3 & 0.7 & \textbf{71.2} & \textbf{23.6} & 18.3 & 6.2 & 12.1 & 0.7 \\
        DRANet-SWD & \textbf{33.77} & \textbf{81.1} & \textbf{33.4} & 74.0 & 17.2 & 15.0 & 22.0 & 22.1 & 13.8 & 79.4 & 26.9 & 76.2 & \textbf{47.7} & 1.8 & 67.6 & 20.0 & 13.4 & 14.7 & \textbf{14.7} & 0.7 \\
        DRANet + Pretrain & 32.62 & 74.8 & 23.3 & \textbf{76.4} & 19.4 & 10.8 & 21.7 & 21.4 & 10.5 & 81.7 & 30.8 & 76.6 & 44.3 & 2.0 & 64.7 & 18.5 & 19.3 & \textbf{16.4} & 5.9 & 1.4 \\
        DRANet-SWD + Pretrain & 33.74 & 59.1 & 22.9 & 74.0 & 16.9 & \textbf{16.3} & 21.9 & \textbf{25.5} & \textbf{16.1} & \textbf{82.6} & \textbf{37.3} & \textbf{77.9} & 47.3 & \textbf{12.6} & 58.1 & 22.9 & \textbf{25.4} & 6.9 & 14.1 & \textbf{3.2} \\
        \bottomrule
    \end{tabular}
    \label{tab:segmenation_results_table}
\end{table*}

\begin{figure*}[htbp]
    \centering
    \begin{subfigure}[b]{0.49\textwidth}
        \centering
        \includegraphics[width=0.95\textwidth]{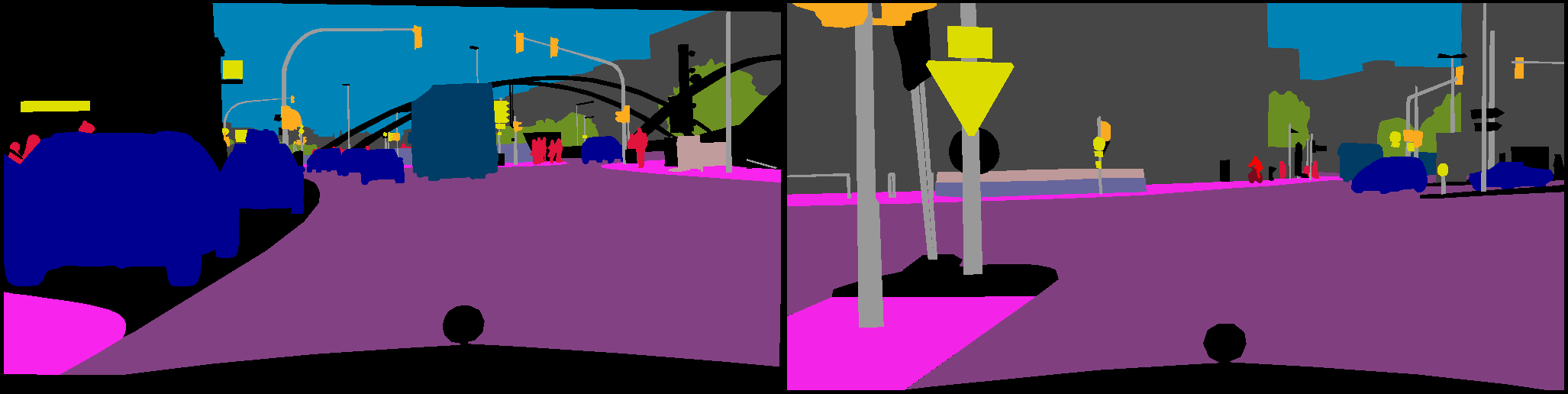}
        \caption{Cityscapes Segmentation Ground Truth}
         \label{fig:seg_a}
    \end{subfigure}
    \begin{subfigure}[b]{0.49\textwidth}
        \centering
        \includegraphics[width=0.95\textwidth]{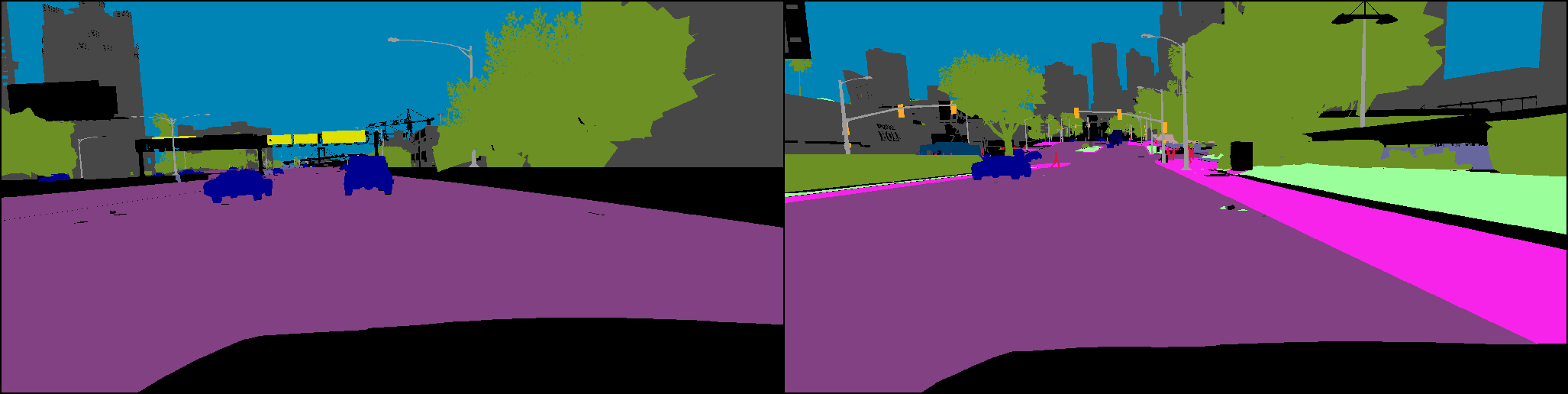}
        \caption{GTA5 Segmenatation Ground Truth}
        \label{fig:seg_b}
    \end{subfigure}
    \begin{subfigure}[b]{0.49\textwidth}
        \centering
        \includegraphics[width=0.95\textwidth]{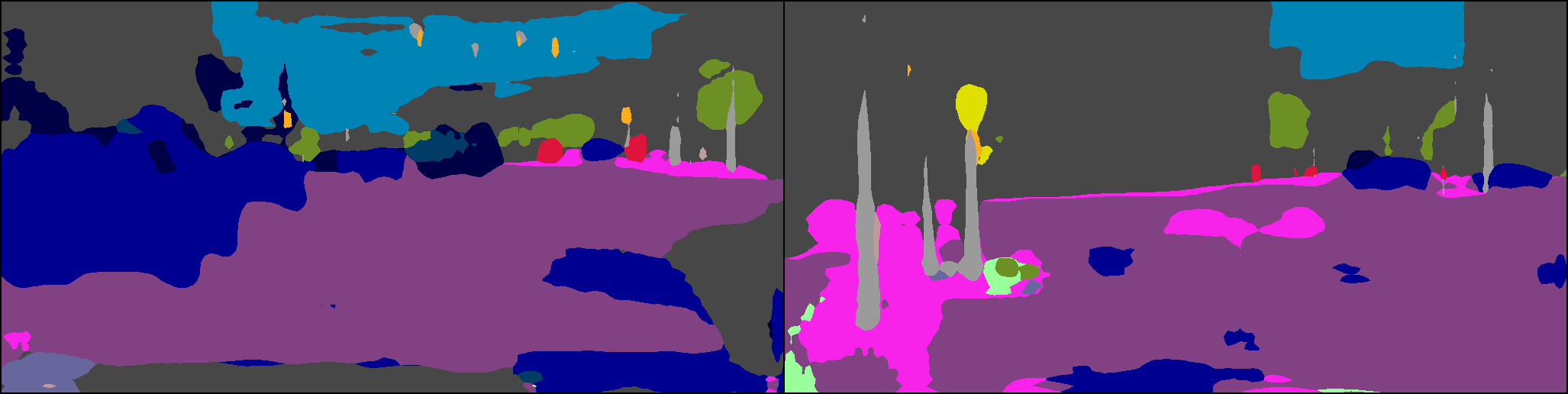}
        \caption{DRANet-SWD prediction Cityscapes}
         \label{fig:seg_c}
    \end{subfigure}
    \begin{subfigure}[b]{0.49\textwidth}
        \centering
        \includegraphics[width=0.95\textwidth]{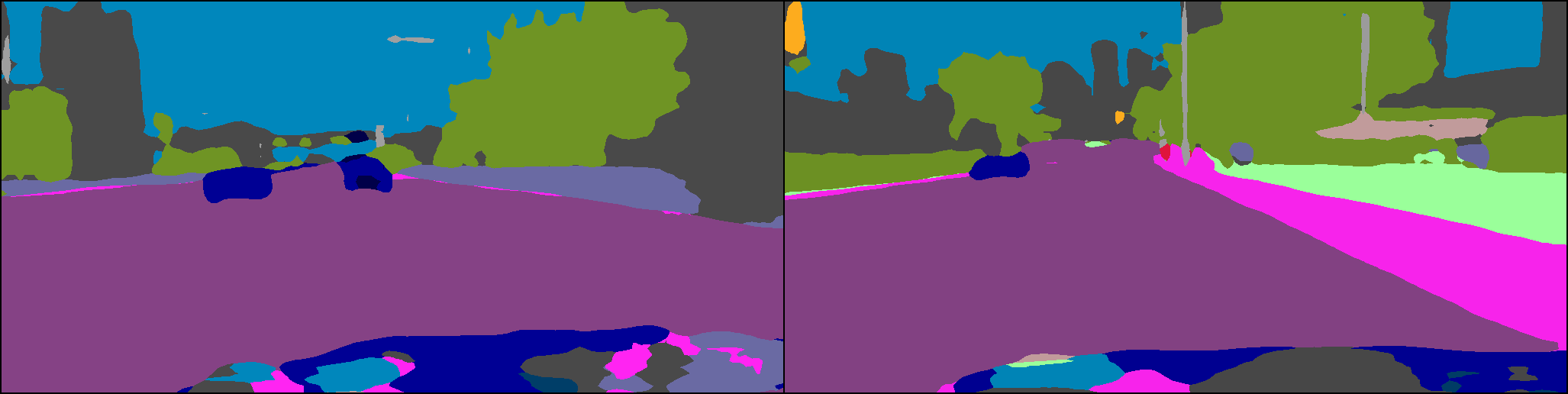}
        \caption{Pretrained DRANet-SWD prediction GTA5 $\rightarrow$ Cityscapes}
         \label{fig:seg_d}
    \end{subfigure}
    \begin{subfigure}[b]{0.49\textwidth}
        \centering
        \includegraphics[width=0.95\textwidth]{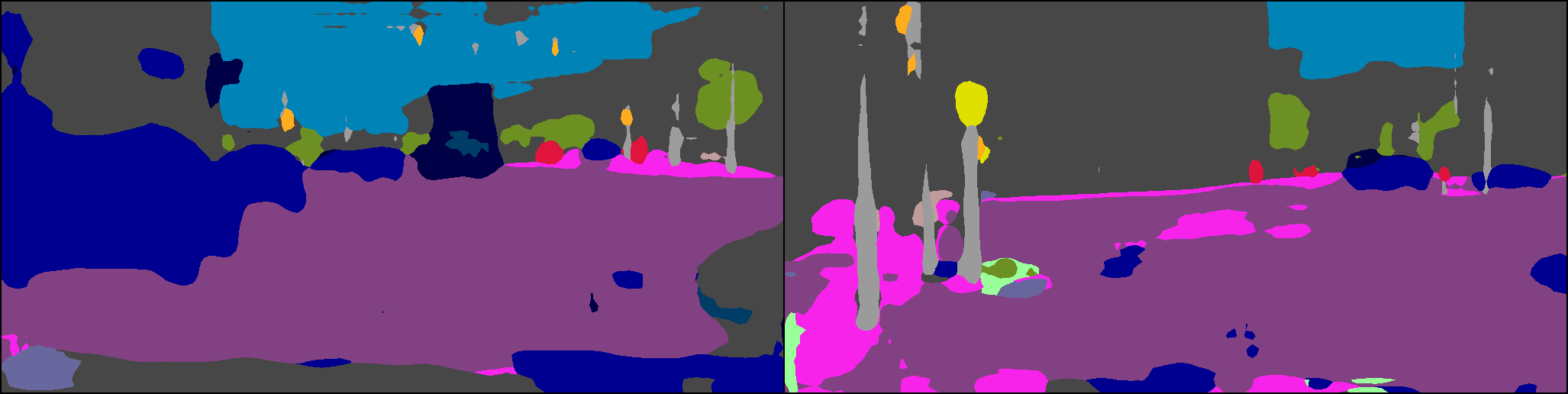}
        \caption{DRANet prediction Cityscapes}
         \label{fig:seg_e}
    \end{subfigure}
    \begin{subfigure}[b]{0.49\textwidth}
        \centering
        \includegraphics[width=0.95\textwidth]{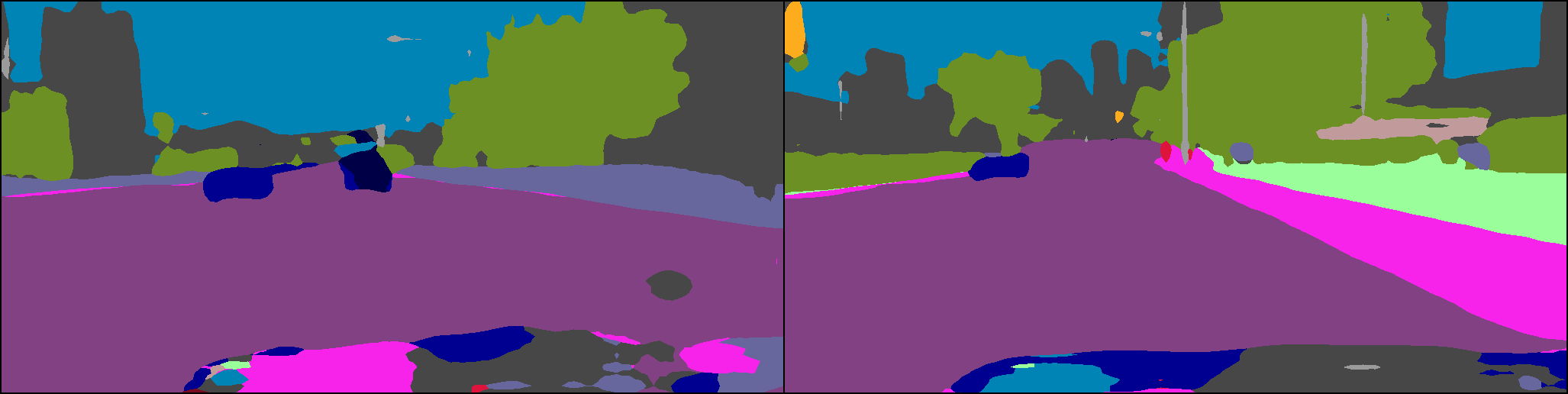}
        \caption{Pretrained DRANet prediction GTA5 $\rightarrow$ Cityscapes}
         \label{fig:seg_f}
    \end{subfigure}
    \caption{GTA5 $\rightarrow$ Cityscapes semantic segmentation results for DRANet and DRANet-SWD.}
    \label{fig:segmentation}
\end{figure*}
The digit classification results are shown in Table \ref{tab:digits_results} and the associated images are shown in Fig \ref{fig:Digit classification}. DRANet-SWD shows an improvement in most adaptations with exceptions being MNIST to MNIST-M and SVHN to MNIST.
DRANet-SWD shows a great improvement for adapting MNIST to USPS and vice versa. These datasets are both grayscale with differing resolutions, USPS being 16x16 and MNIST being 28x28, as well as differing stroke styles and thicknesses. These datasets see a valuable improvement in classification accuracy likely due to SWD more effectively capturing the textural style between images with differing resolutions. SWD functioning with probability distributions may generalize better in this case where standard feature alignment has more difficulty bridging this gap.
DRANet-SWD nearly matches the performance of DRANet for adapting MNIST to MNIST-M and vice versa. It does seem to misinterpret a style rarely for MNIST to MNIST-M but most styles seem to have improved fidelity compared to DRANet. The content of the images produced by DRANet-SWD seem much clearer especially for MNIST-M to MNIST with several examples showing DRANet-SWD succeeds where DRANet fails.
The domain transfer tasks involving SVHN show the limitations of the content and style representation approach. Other latent space approaches for this specific dataset such as triplet loss networks \cite{triplet_loss_network} have superior performance. In the SVHN dataset, multiple numbers are present confusing content and style representations which severely impacts performance. Interestingly when examining Fig \ref{fig:digit_c} DRANet-SWD has partially and poorly adapted some digits, while DRANet was unable. In order to achieve the results shown in Table \ref{tab:digits_results} for this MNIST to SVHN and vice versa, a more expressive classifier was required than was used for the other digit classification tasks.

To ensure reproducibility, we conducted each experiment using multiple seeds. The Github repository for DRANet uses 5688 as a chosen seed. For the adaptation between MNIST and USPS, we use additional seeds 292 and 3445, while for the adaptation between MNIST and MNIST-M, we use additional seeds 9901, and 2516. The original DRANet paper doesn't showcase MNIST to SVHN results so the 3 random seeds were 8797, 8826, and 9330. Reported results in Table \ref{tab:digits_results} reflect the mean performance and standard deviation across these runs.

\begin{table}[pbht]
\centering
\caption{Unsupervised domain adaptation digit classification results comparing DRANet-SWD and DRANet for 3 random seeds trained to 40000 iterations.}
\begin{tabular}{lc|c|c}
\multirow{2}{*}{\textbf{Domains}} & \multicolumn{2}{c}{\textbf{Models}} \\
  & DRANet (Gram) & DRANet-SWD \\ \hline
{\textit{MNIST to UPS}} & 94.98\%$\pm$\textbf{0.23\%} & \textbf{95.28\%}$\pm$0.47\%\\ \hline
{\textit{UPS to MNIST}} & 92.03\%$\pm$0.78\% & \textbf{93.67\%}$\pm$\textbf{0.42\%}\\ \hline
{\textit{MNIST to MNIST-M}} & \textbf{95.53\%}$\pm$0.44\% & 95.25\%$\pm$0.44\%\\ \hline
{\textit{MNIST-M to MNIST}} & 98.62\%$\pm$\textbf{0.07\%} & \textbf{98.68\%}$\pm$0.17\%\\ \hline
{\textit{MNIST to SVHN}} & 19.8\%$\pm$\textbf{2.7\%} & \textbf{47.3\%}$\pm$4.5\%\\ \hline
{\textit{SVHN to MNIST}} & \textbf{60.0\%}$\pm$10.3\% & 36.1\%$\pm$\textbf{6.4\%}\\ \hline
\end{tabular}
\label{tab:digits_results}
\end{table}

\subsection{Driving Scene Segmentation}
The results for the GTA5 to Cityscapes segmentation are shown in Table \ref{tab:segmenation_results_table} and the segmented images shown in Fig \ref{fig:segmentation}. The results compare DRANet and DRANet-SWD with and without the further training for the segmentation network. Overall DRANet-SWD marginally outperforms DRANet achieving slightly higher mean Intersection-over-Union (mIoU) and demonstrates higher accuracy in several categories. Interestingly, DRANet achieves better accuracy in several categories such as road, sidewalk, car and truck without pretraining. However, DRANet-SWD shows an improvement when classifying smaller or less frequently occurring objects, suggesting that the SWD mechanism improves robustness to rare categories. Its important to note that these results under perform in comparison to the results reported by Lee et al.\cite{DRANet_CVPR2021} suggesting differing training setups. An avenue for further experimentation include additional hyperparameter tuning including batch size, SWD projection number etc to ensure similar results as previously reported.

These results were obtained using a fixed random seed (5688), as provided in the DRANet GitHub repository.

\section{Conclusions}
In this paper, we introduced DRANet-SWD, a disentangled representation learning framework for unsupervised domain adaptation that replaces the traditional Gram matrix style loss with sliced Wasserstein discrepancy (SWD). While our independent implementation of DRANet did not achieve the originally reported results, the incorporation of SWD into this method in the proposed pipeline consistently improved results. Specifically, our model better captures the underlying statistical properties of style representations by leveraging SWD, leading to domain adaptation performance that is better or equivalent to the existing method. Our experiments demonstrate that DRANet-SWD enhances visual fidelity and adaptation effectiveness across both classification and segmentation tasks, highlighting the advantages of SWD over traditional style transfer methods. We also discuss and showcase the datasets where the model struggles. These results reinforce the importance of distribution-based metrics in representation learning and provide a strong foundation for further advancements in domain adaptation. Additional avenues for improvement include analyzing existing or developing new representation strategies for adapting difficult problems such as adapting MNIST and SVHN unsupervised.

{
\bibliographystyle{IEEEtran}
\bibliography{ref}
}

\end{document}